\documentclass[conference]{IEEEtran}
\IEEEoverridecommandlockouts
\usepackage{cite}
\usepackage{amsmath,amssymb,amsfonts}
\usepackage{algorithmic}
\usepackage{graphicx}
\usepackage{textcomp}
\usepackage{xcolor}
\def\BibTeX{{\rm B\kern-.05em{\sc i\kern-.025em b}\kern-.08em
    T\kern-.1667em\lower.7ex\hbox{E}\kern-.125emX}}
\usepackage{times}
\usepackage{latexsym}
\usepackage{subcaption}
\DeclareMathOperator*{\argmax}{arg\,max}

\usepackage{url}
\usepackage{color}

\usepackage{mathtools}
\DeclarePairedDelimiter\abs{\lvert}{\rvert}

\definecolor{b1}{RGB}{8,81,156}
\definecolor{b2}{RGB}{49,130,189}
\definecolor{b3}{RGB}{107,174,214}
\definecolor{b4}{RGB}{158,202,225}

\begin{document}

\title{Comprehensible Context-driven Text Game Playing}

\author{\IEEEauthorblockN{Xusen Yin}
\IEEEauthorblockA{\textit{Information Sciences Institute} \\
\textit{University of Southern California}\\
Marina del Rey, CA\\
\texttt{xusenyin@usc.edu}}
\and
\IEEEauthorblockN{Jonathan May}
\IEEEauthorblockA{\textit{Information Sciences Institute} \\
\textit{University of Southern California}\\
Marina del Rey, CA \\
\texttt{jonmay@isi.edu}}
}

\maketitle

\begin{abstract}
In order to train a computer agent to play a text-based computer game, we must represent each hidden state of the game. A Long Short-Term Memory (LSTM) model running over observed texts is a common choice for state construction. However, a normal Deep Q-learning Network (DQN) for such an agent requires millions of steps of training or more to converge. As such, an LSTM-based DQN can take tens of days to finish the training process. Though we can use a Convolutional Neural Network (CNN) as a text-encoder to construct states much faster than the LSTM, doing so without an understanding of the syntactic context of the words being analyzed can slow convergence. In this paper, we use a fast CNN to encode position- and syntax-oriented structures extracted from observed texts as states. We additionally augment the reward signal in a universal and practical manner. Together, we show that our improvements can not only speed up the process by one order of magnitude but also learn a superior agent.
\end{abstract}

\section{Introduction}

Ever since the work of \cite{google-atari} in learning to play Atari games, the question of whether we could also learn to play text-based games has naturally arisen \cite{fulda2017affordance,DBLP:conf/nips/ZahavyHMMM18,textworld-a-learning-environment-for-text-based-games}. However, the goal of reaching par with human players on these games is still beyond our reach, unlike that shown for Atari games. Text-based games, especially those designed for real human players, are elaborately built and hence sophisticated. Zork \cite{zork,textworld-a-learning-environment-for-text-based-games} is one such game, with more than 30 rooms to explore, and combines a maze, trivia, combat, time-sensitive tasks, puzzles, and stochastic events. Most attempts to automatically learn to play real text games can only explore a few rooms of a game, achieving about 10 percent of the total available score.

Unlike Atari action games where one uses the joystick to play, and thus has up to 18 different actions available (including a button press), a player uses brief natural language sentences as actions to interact with text-based games. The number of valid actions is thus theoretically infinite, and even when the vocabulary and maximum action length are limited, can still be in the hundreds or even thousands. This makes policy-based learning quite difficult.

The Deep Q-learning Network (DQN), first introduced by \cite{google-atari}, is also the main method used to play text-based games. One key component of the DQN as applied to text games is its encoding of context sentences into hidden states to represent game state. As the length of context sentences can vary from a few words to thousands of words, the LSTM \cite{lstm-original} is seemingly a natural choice for this task \cite{DBLP:conf/cig/KostkaKKR17,D15-1001,DBLP:journals/corr/abs-1805-07274,textworld-a-learning-environment-for-text-based-games}.

However, one important limitation of DQN learning is that it usually needs millions of training steps to converge, and this can take days with the LSTM as the encoder, even for a small quest with only tens of actions. Since full-fledged text-based games usually involve multiple quests and have many more actions available to try, training an LSTM-based DQN agent for text-based games such that it may reach a level of quality  similar to that observed for  Atari games is functionally impossible.

\cite{D14-1181} shows that we can use a convolutional neural network (CNN) as a text encoder in classification tasks for faster training. \cite{DBLP:conf/nips/ZahavyHMMM18} uses a similar CNN encoder to build a DQN agent. However, previous work of building DQN agents focuses on DQN architecture \cite{D15-1001,P16-1153,DBLP:journals/corr/abs-1805-07274} or the generation and selection of actions \cite{fulda2017affordance}, but lacks much analysis of the observed context sentences. 

In this paper, we focus on the analysis of observed texts and how they work with different encoder architectures. We show that both context sentence processing and encoder selection can lead to faster convergence of the DQN training process and result in superior agents. 
After analyzing the training process, we also find that instant reward manipulation and sample strategies can affect the training process. Furthermore, we observe that the CNN encoder with a max-pooling layer can be treated as an auto-attention mechanism in finding key components in the context to make decisions. Our final trained agent on Zork can reach state-of-the-art scores within one million training steps, or about 10 hours.\footnote{Our code is available at \url{https://github.com/yinxusen/dqn-zork}}

The novel contributions of our paper are: 
\begin{itemize}
    \item We compare different encoders for the DQN framework, and determine that using CNN with a max-pooling layer as the encoder is ideal, and has the extra benefit that max-pooling functions as an  auto-attention mechanism. 
    \item We use position embeddings when encoding trajectories to reach the state-of-the-art result on Zork;
    \item We use a dependency parser to reorder game context sentences such that syntactically related elements are close to each other, which leads to a halving of  our DQN agent's convergence time. We use reward shaping methods for further better training results.
\end{itemize}

\section{Methods}

\subsection{Text-based games}\label{sec:tbg-egg-example}

When playing a text-based game, the game first outputs a sentence to describe the current environment. Then a player inputs a sentence as the action to the game, and waits for the next game output and a cumulative score from the beginning of the game. An example of taking an egg in Zork is
    
\begin{small}
{\it $l_1$: West of House
You are standing in an open field west of a white house, with a boarded
front door.
There is a small mailbox here. (Score: 0, Moves: 0)}

{\it $l_2$: $>$ go north}

{\it $l_3$: North of House
You are facing the north side of a white house. There is no door here, and
all the windows are boarded up. To the north a narrow path winds through
the trees. (Score: 0, Moves: 1)}

{\it $l_4$: $>$ go north}

{\it $l_5$: Forest Path
This is a path winding through a dimly lit forest. The path heads north-south here. One particularly large tree with some low branches stands at
the edge of the path. (Score: 0 Moves: 2)}

{\it $l_6$: $>$ climb tree}

{\it $l_7$: Up a Tree
You are about 10 feet above the ground nestled among some large branches.
The nearest branch above you is above your reach.
Beside you on the branch is a small bird's nest.
In the bird's nest is a large egg encrusted with precious jewels,
apparently scavenged by a childless songbird. The egg is covered with fine
gold inlay, and ornamented in lapis lazuli and mother-of-pearl. Unlike most
eggs, this one is hinged and closed with a delicate looking clasp. The egg
appears extremely fragile. (Score: 0, Moves: 3)}

{\it $l_8$: $>$ take the egg}

{\it $l_9$: Taken. (Score: 5, Moves: 4)}
\end{small}

We call the game output the {\it master}, a player's input sentence the {\it action}, and the gap between two consecutive scores the {\it instant reward}. A running log of a text-based game is a composition of master-action pairs, which we call a {\it trajectory}. In this example, $l_1$, $l_3$, $l_5$, and $l_7$ each are masters, while $l_2$, $l_4$, $l_6$, and $l_8$ each are actions. The instant reward of taking the egg is five points. The trajectory using to make the decision of choosing {\it take the egg} is the concatenation from $l_1$ to $l_5$, to denote this we use the label $t_{1-5}$.

\subsection{DQN framework}



A text-based game constitutes a set of states $S$, a set of actions $A$, a transition matrix $T: (S, A) \to S$, and an instant reward matrix $R: (S, A) \to \mathcal{R}$. At a state $s$, the game accepts an action $a$, then transitions to new state $s'$ with master $m$ and instant reward $r$. The game terminates upon reaching certain terminal states, e.g. the actor is defeated in fighting. If $S$ and $A$ are finite, we can search for a policy to play the game with (simple) Q-learning: 
Let $Q: (S, A) \to \mathcal{R}$ be a matrix over the (state, action) space $(S, A)$. Each value $Q[s, a]$ is the {\it expected reward} that the agent will get in the future until termination if choosing $a$ at state $s$. Given the stochastic attribute of the FST, a transition of $T[s, a]$ could result in multiple new states. E.g. Using the same action $a$ at the state $s$, with probability of 0.7 the player enters a room with no thief ($s'_{1}$), while with probability of 0.3 the player enters the same room by with a thief in there ($s'_{2}$). Computing the Q-value of the state $s$ needs to compute the weighted sum of both of the next states $s'_{1}$ and $s'_{2}$. We then have an iterative update function for computing $Q$:

\begin{equation} \label{eq:dqn-update}
Q[s, a] = E_{T[s, a]}\left(R[s, a] + \gamma * \max_{a'}(Q[T[s, a], a']])\right),
\end{equation}

\noindent where $E_{T[s,a]}$ is the expectation according to multiple new states, and $\gamma$ is a hyperparameter that controls the importance of future rewards.

With the knowledge of the Q-matrix, the policy of playing a game is to choose the action with the highest Q-value at each state $s$,

\[a \leftarrow \argmax Q[s, *].\]

The DQN was introduced by \cite{google-atari} for playing Atari video games. In video games, the FST's state is implicitly provided by video frames, while the action is explicitly encoded as a joystick movement combined with the presence or absence of a button press. \cite{google-atari} use a CNN to encode video frames into compact hidden states. The Q-matrix is then represented by a function $f_{DQN}: (S, A) \to \mathcal{R}$. Unlike simple Q-learning, deep Q-learning allows for very large or even infinite $A$ and $S$.

According to the Q-matrix update in Equation \ref{eq:dqn-update}, DQN training requires training samples in the form of a tuple $(s, s', a, r)$ consisting of the current state $s$, the action $a$, the next state $s'=T[s, a]$, and the instant reward $r=R[s,a]$. In text-based games, the state of the game is not simply a game location such as \textit{up a tree}, but is a trajectory. In our egg taking example, at both $l_7$ and $l_9$ the player is in the location {\it up a tree}, but in $l_9$ the player also has an egg in the inventory. Having the complete trajectory is a means for easily differentiating these states. 

With our egg taking example,  four training samples exist:

 {\it ($t_{1-1}$, $t_{1-3}$, go north, 0)},
 
  {\it ($t_{1-3}$, $t_{1-5}$, go north, 0)},

 {\it ($t_{1-5}$, $t_{1-7}$, climb tree, 0)},

 {\it ($t_{1-7}$, $t_{1-9},$ take the egg, 5)}.

The DQN training is a process of repeatedly playing the game in an exploration-exploitation manner: At every step, the DQN agent chooses a random action to play the game with a probability $\epsilon$ (exploration), or else it chooses the action with the best Q-value (exploitation). The $\epsilon$ decays from 1 to almost zero to anneal this process.

The DQN collects the training sample $(s, s', a, r)$ at every step into a  \textit{replay memory}. The training process gets samples from the replay memory to update the DQN. The loss function to update the DQN is  

\[\left(f_{DQN}(s, a) - (r + \gamma * \max_{a'} f_{DQN}(s', a')\right)^2,\]

\noindent which is the square error loss between expected Q values and predicted Q values.

We use the same DQN framework to play text-based games, but encode texts instead of video frames.

\subsection{Trajectory Encoder}

\begin{figure}
    \centering
    \includegraphics[width=0.45\textwidth]{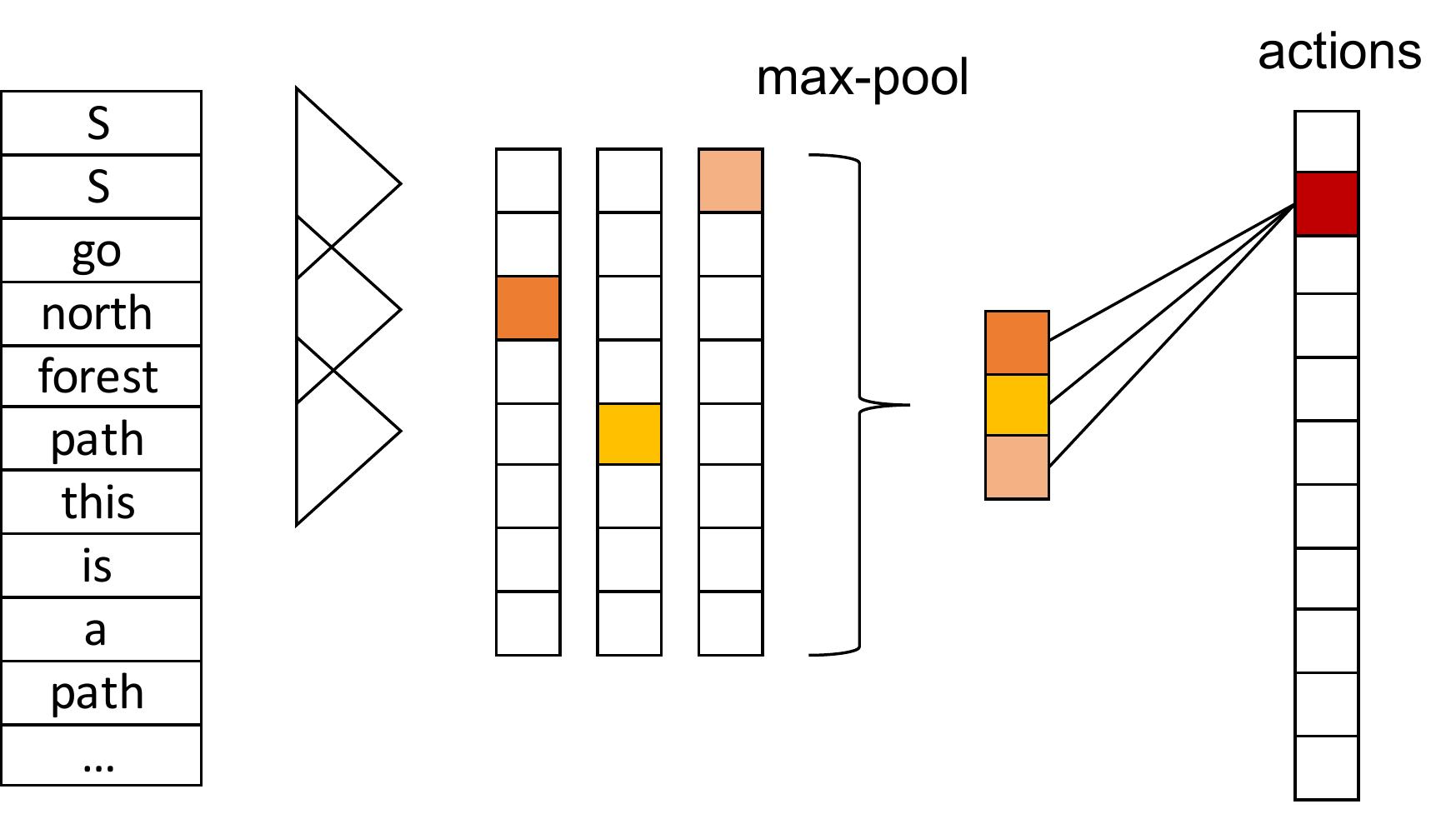}
    \caption{An example of CNN-DQN using three size-3 convolutional filters with a max-pooling layer on a trajectory {\it go north forest path this is a path ...}. {\it S} is the start of sentence token. We use two start of sentence tokens to make sure the vectors generated from convolutional filters have the same length with the original trajectory. Each convolutional filter (triangle) encoding the whole trajectory along the dimension of tokens generates a feature vector.}
    \label{fig:cnn-max-pooling-arch}
\end{figure}

The purpose of the trajectory encoder is to determine the hidden states of the game. We initially use an LSTM \cite{lstm-original} as the encoder. The LSTM encodes sentences in a recurrent way, with an inner state vector to be updated after consuming each token, yielding two types of output: an output vector for each consumed token, and a final state. \cite{D15-1001} uses the result vector from a mean-pooling layer on the output vector as the state, while we use final state as the encoded state. Our CNN encoder is inspired by \cite{D14-1181}, but we also use different pooling layers. The CNN encoder uses multiple one-dimensional convolutional filters with different kernel sizes to encode sentences, then uses a mean-pooling layer or a max-pooling layer along the dimension of the sentence, and finally concatenates pooling results into a one-dimensional vector. An example of the max-pooling CNN-DQN is shown in Figure \ref{fig:cnn-max-pooling-arch}.

The fact that the CNN encoder encodes blocks of tokens in a parallel way makes it much faster than the token-by-token LSTM encoder. However, the vanilla CNN encoder loses track of the position relationship between tokens, since it treats sentences as bags of words. To mitigate this, we apply position embeddings together with word embeddings to keep the position information, in the same way as \cite{pmlr-v70-gehring17a}. As we will see in experiments, the CNN encoder with the use of position embeddings results in faster training and superior agents.

\section{Experiment setup}
\label{sec:expsetup}

We experiment using the classic 1979 game Zork I \cite{zork}. 
All experiments are run on a machine with two Tesla K80 GPUs (one for training, the other for evaluation) with four CPU threads. We use Python 3.6.8 and Tensorflow 1.12.0.


A game log collected from an expert player \cite{zork-transcription} to win all points (350) of Zork I consists of 345 steps with 130 types of actions. We think of the log as a near-optimal solution to solve Zork and constrain our system to use these 130 actions in our experiments.

\subsection{Reward shaping and clipping}

Reward shaping can embed common sense into the training process, leading to better agents. 
For the sake of faster convergence, we add $-0.1$ to all instant rewards. Negative masters such as ``{\it you don't ...}'' and ``{\it you can't ...}'' mean there is something wrong with our chosen actions, so we add a penalty of $-1$ on instant rewards.

We find that reward clipping is important in training the DQN to play text-based games. \cite{D15-1001} uses a reward shaping method to make the expected state more noticeable by magnifying quest final scores. In this paper, we use a similar reward clipping method as \cite{google-atari}. We impose a variant of Huber loss \cite{huber1964}, clipping the reward so it falls between -1 and 1, without modification of rewards that are initially in that range. Clipping rewards in this way is robust to outliers and independent of the reward systems introduced by different games, resulting in better generalization ability to all other games.

\subsection{Hyperparameters}

Careful choice of hyperparameters is vital to successful stochastic DQN training. We use 50,000 observation steps, 500,000 replay memories, and linearly decay $\epsilon$ from 1 to 0.0001 in 2,000,000 steps. Since the near-optimal path to solving Zork is 345 steps, we set each episode to have a maximum of 600 steps.

We save a DQN model and run evaluation every 5,000 steps of training (as one epoch). For each evaluation, we use a fixed $\epsilon=0.05$ as other work did \cite{google-atari, fulda2017affordance, D15-1001} and run 10 episodes with the same number of steps per episodes for training.

We initialize our DQN models with random word embeddings and position embeddings. We use a fixed embedding size of 64. At every training step, we draw a minibatch of 32 samples and use a learning rate of $1e-5$ with the Adam optimizer. We trim trajectories to contain no more than 21 sentences to avoid unnecessarily long concatenated strings.

\section{Results and Discussion}

\subsection{Uniform sampling, weight-, and gap-based sampling}

Uniform sampling, as first introduced in \cite{google-atari}, is the most common method to draw samples from the replay memory to train the DQN. However, it is not a general method that could work well from game to game.
From our observation, the distribution of rewards is highly biased in the replay memory: Samples with negative and zero rewards are more likely to appear while those with positive rewards are rare.

In an experiment using only 11 actions, where we collect 43,069 training samples, 98.7\% of them receive negative or zero rewards, while only 1.3\% of them are positive rewards. For another experiment using 21 actions with 58,608 samples, 99.9\% of them are negative or zero samples, and only 0.1\% of them are positive ones. Training a DQN agent with uniform sampling on these replay memories can easily miss important samples with positive rewards and lead to a failed agent.

Since uniform sampling will naturally lead to an imbalance favoring zero or negative rewards, we seek a non-uniform sampling method during DQN training. We explore two different weighted sampling strategies in our experiments, \textit{fixed-weight} and \textit{priority experience} sampling. Fixed-weight sampling is based on the previous observation: {\it positive reward samples are rare but more important}. Even though we could use frequency counting as weights, a reward-based weight is more general, and is amenable to scenarios where counting frequency is difficult. We treat the instant reward $r_i$ for the $i$-th sample as the log of weight for sampling, i.e. we let:

\[w_i = exp(r_i), \]

\noindent then the probability of choosing sample $i$ is 

\[P(i) = \frac{w_i}{\sum_j w_j}.\]

Reward clipping needs to be used to avoid samples with very large rewards suppressing all other samples.

Priority experience sampling \cite{DBLP:journals/corr/SchaulQAS15} is based on the gap $\abs{\delta_i}$ between the expected Q-value and the predicted Q-value, i.e. we let:

\[\delta_i = f_{DQN}(s_i, a_i) - \left(r_i + \gamma * \max_{a'} f_{DQN}(s_i', a')\right),\]

\noindent then

\[w_i = \abs{\delta_i} + e,\]

\noindent where $e$ is a small constant to avoid zero gap. The probability of choosing sample $i$ is

\[P(i) = \frac{w_i^a}{\sum_j w_j^a}.\]

The hyperparameter $a$ is a choice of randomness of choosing samples. Setting $a=0$ degrades the sampling method to uniform sampling.

To avoid bias towards samples with high probabilities, the priority experience sampling uses importance weights on the gradient $g$ when updating parameters, i.e.

\[g_i = g_i * \left(\frac{1}{N*P(i)}\right)^b.\]

The hyperparameter $b$ is annealed from 0 to 1 during training.

We will compare the weighted sampling and the priority experience sampling in later sections.

\begin{figure}
    \centering
    \includegraphics[width=0.45\textwidth]{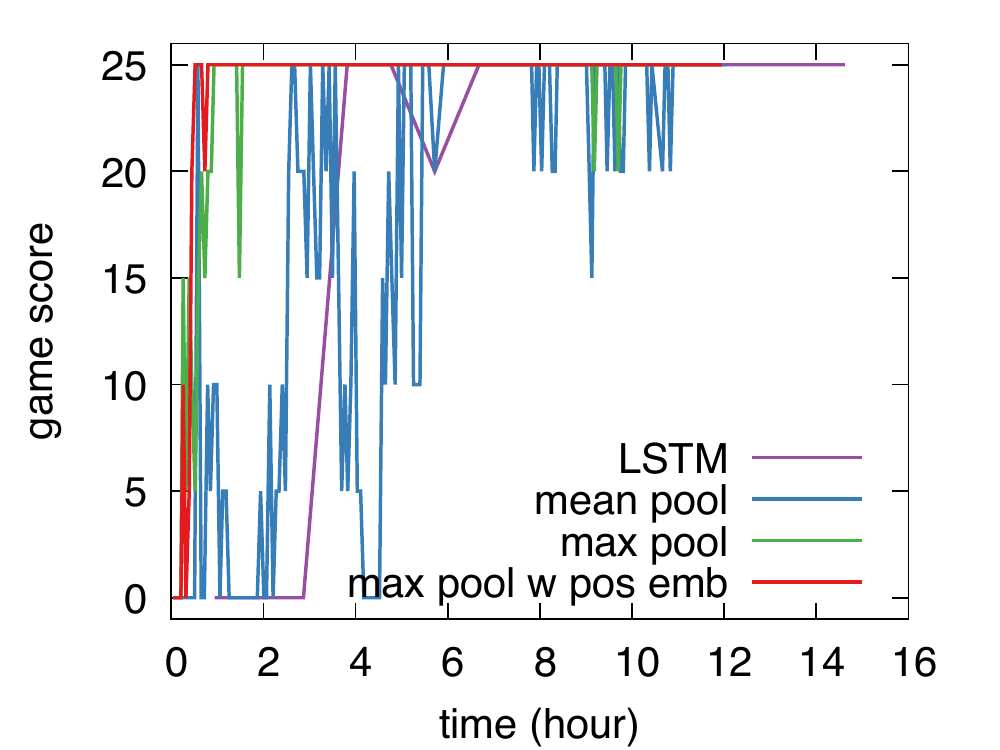}
    \caption{Evaluation results of LSTM-DQN, CNN-DQN, and CNN-DQN with position embeddings on the egg quest of Zork I (as defined in Section~\ref{sec:expsetup}). In this plot, CNN-based DQNs are trained around 175 epochs, while the LSTM-DQN is trained 14 epochs in the same amount of time. All DQNs converge to reach a score of 25 at the end of training except the mean-pooling CNN-DQN (blue) that jitters even after 10 hours of training. On the contratry, the ones using a max-pooling layer (red and green) show more stable convergence curves, and the red curve converges fastest in half an hour using position embeddings. The LSTM-DQN (purple) converges after 7 hours, slower than the max-pooling CNN-DQNs.}
    \label{fig:egg-quest}
\end{figure}

\begin{figure}
    \centering
    \includegraphics[width=0.45\textwidth]{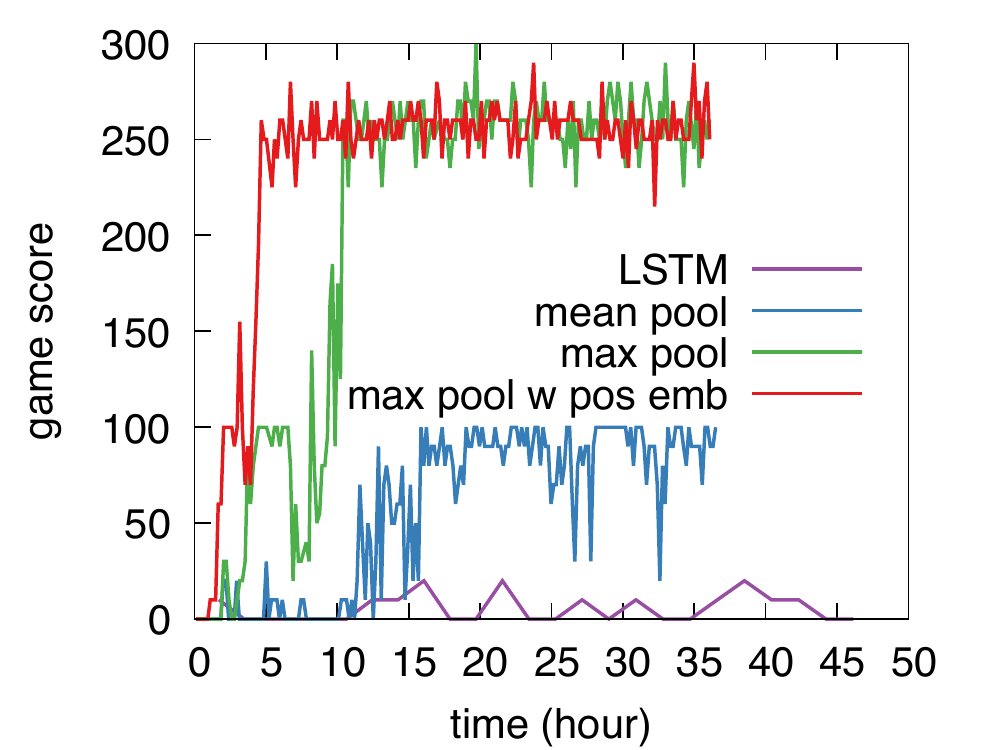}
    \caption{Evaluation results of LSTM-DQN, CNN-DQN, and CNN-DQN with position embeddings on the troll quest of Zork I. In this plot, CNN-DQNs are trained for 190 epochs, while the LSTM-DQN is trained 25 epochs in the same amount of time. The max-pooling CNN-DQNs (red and green) converge fast with higher scores than the others, and the red one (using position embeddings) is the fastest to converge, at 5 hours. The mean-pooling CNN-DQN (blue) shows convergence at a lower score but with more jitters. The LSTM-DQN (purple) cannot be well-trained in 45 hours of training.}
    \label{fig:troll-quest}
\end{figure}

\begin{figure}
    \centering
    \includegraphics[width=0.45\textwidth]{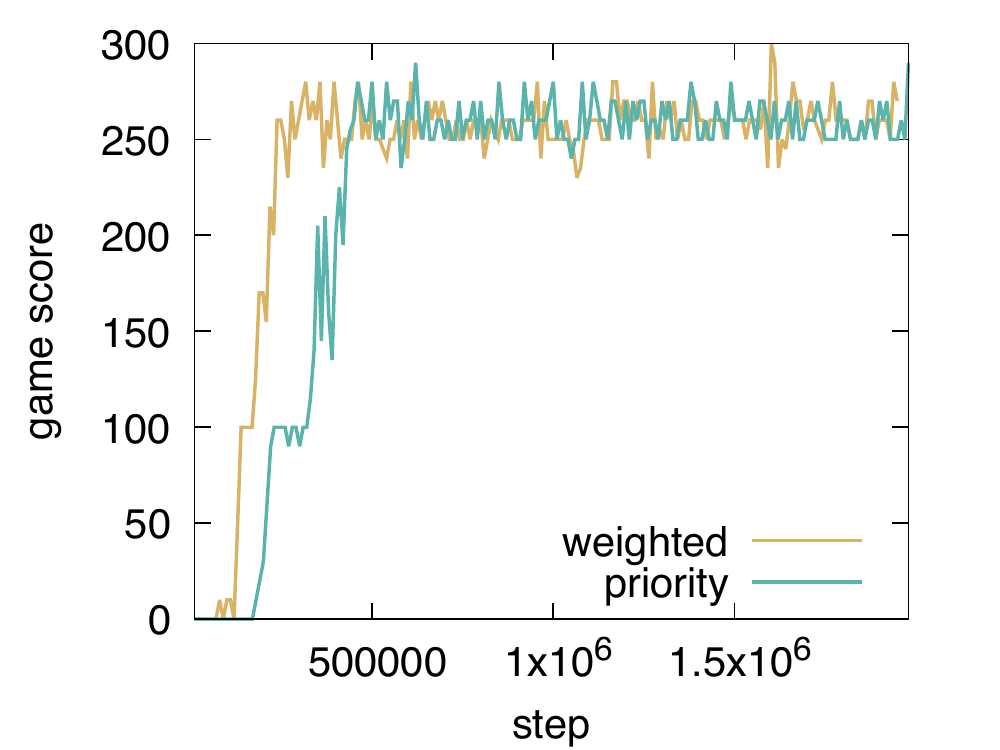}
    \caption{Evaluation results of the troll quest with the max-pooling CNN-DQN. The yellow curve uses weighted sampling while the green one uses priority experience sampling. The weighted sampling method leads to faster convergence.}
    \label{fig:weighted-vs-pes-troll}
\end{figure}

\subsection{Choosing the encoder}

Since we use the exploration-exploitation search method with a decaying parameter $\epsilon$, it is better to let the $\epsilon$ decay to almost zero to see a whole picture of the training process. However, a well-trained LSTM-DQN on Zork could take tens of days. In order to compare encoders faster, we consider two of the sub-tasks that Zork comprises, following the approach taken by  \cite{DBLP:conf/nips/ZahavyHMMM18}.

{\bf Egg quest:} From the beginning of the game, the player is required to walk to a tree, climb on the tree, and take an egg. The optimal action combination to win the task is in four steps as shown in our egg-taking example in Section \ref{sec:tbg-egg-example}. Zork gives the player 5 points for getting the egg. To simplify the task, we provide 11 complete actions for the agent to choose from: eight navigation actions and three essential actions to finish the task.

{\bf Troll quest:} From the beginning of the game, the player is required to walk into a house, take a lantern and weapons, find a secret path to a troll room, and kill the troll. This task is more complicated than the egg quest, and one error in the action combination could lead to failure. Also, the player may (stochastically) be killed by the troll, which necessitates a game restart. The optimal action combination is {\it ``go north $|$ go east $|$ open window $|$ enter house $|$ go west $|$ take sword $|$ take lantern $|$ move rug $|$ open trap door $|$ turn on lantern $|$ go down $|$ go north $|$ kill troll with sword.''} We select 21 complete actions for this task. Among the 21 actions, there are eight navigation actions.

Even though we can use the same set of hyperparameters for the egg quest and the troll quest that we use on the complete game, we choose to decrease these hyperparameters for a faster experimental cycle, based on our knowledge that these two quests are sub-tasks extracted from Zork:

\begin{itemize}
    \item For the egg quest we use 5,000 observation steps, 50,000 replay memory size, and we decay $\epsilon$ from 1 to 0.0001 in 500,000 steps. Since the optimal path to solving the egg quest is four steps, we set each episode to have a maximum of 100 steps.
    \item Since the troll quest has a larger search space than the egg quest, we double the observation steps, replay memory size, and $\epsilon$ decaying steps settings used in the egg quest. The optimal path to solving the troll quest is 13 steps, so we set each episode to have a maximum of 150 steps.
\end{itemize}

We compare encoders based on the egg quest and the troll quest. As shown in Figure \ref{fig:egg-quest} and Figure \ref{fig:troll-quest}, we compare four types of encoders: the LSTM encoder, the CNN encoder with mean-pooling, the CNN-encoder with max-pooling, and the use of position embeddings on trajectories. Both figures show consistent results toward these four types of DQNs: 

The LSTM-DQNs (purple) run many fewer epochs than CNN-DQNs in the same amount of time. While LSTMs can converge in the egg quest with 7 hours, they cannot be well-trained in the troll quest within 45 hours, resulting in poor results in the troll quest.

The mean-pooling CNN-DQNs (blue) converge with jitters, and show an inferior result in the troll quest.

The max-pooling CNN-DQNs (red and green) show better results in both their time to   convergence and their maximum scores than other approaches. CNN-DQNs using position embeddings (red) show the best results on both quests.

We compare weighted sampling and the priority experience sampling on the troll quest, as shown in Figure~\ref{fig:weighted-vs-pes-troll}. Both sampling methods converge to about the same score, but weighted sampling converges more quickly. Weighted sampling is suitable for single-quest games that focus on important positive rewards.

For subsequent experiments on the complete Zork, we use the max-pooling CNN-DQN.

\subsection{Pooling layer and auto-attention}

\begin{figure*}
    \centering
    \includegraphics[width=0.9\textwidth]{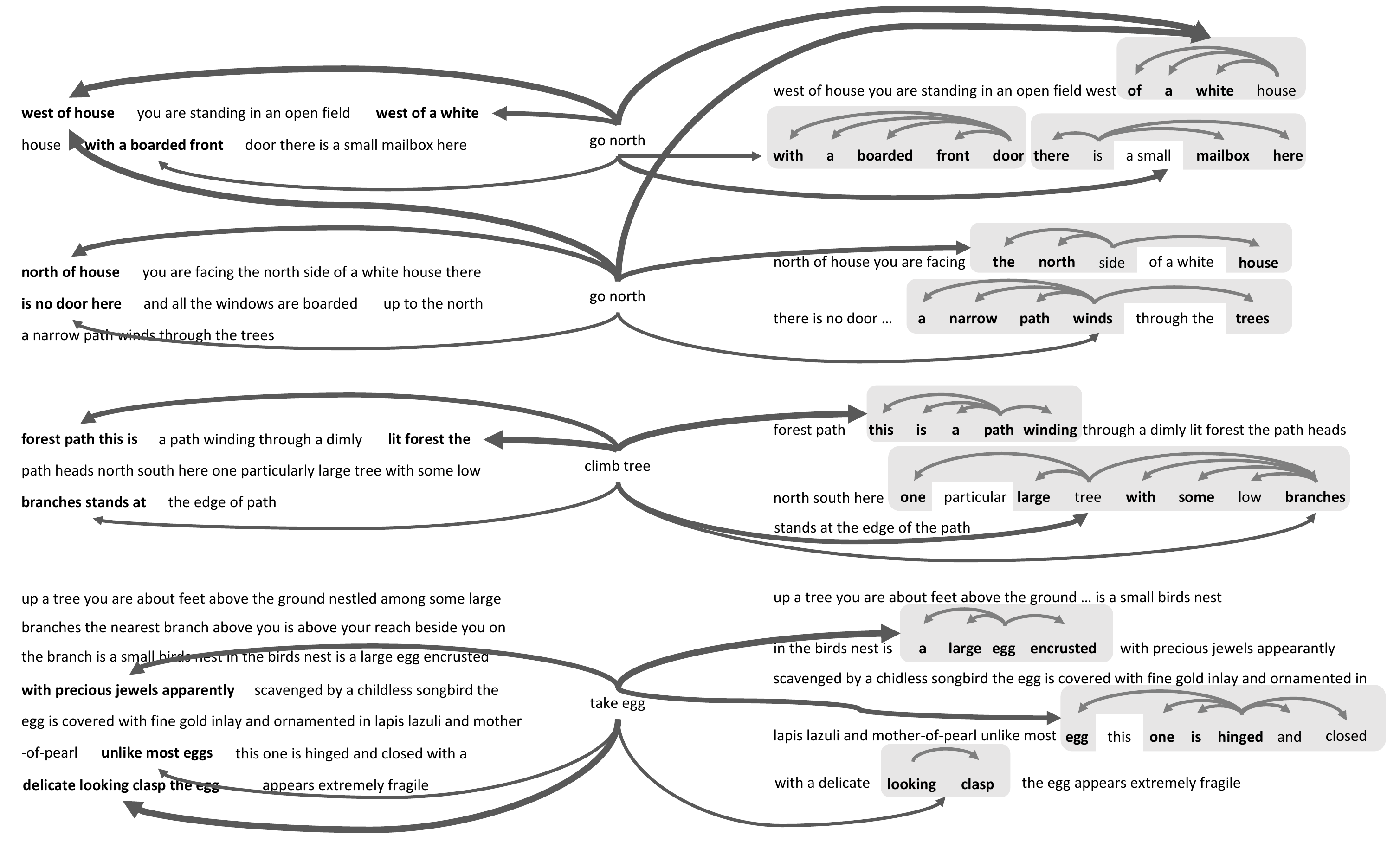}
    \caption{Pseudo-attention via max pooling. For each action chosen (center column), we show the three most important context spans relevant to this decision, both with unmodified context (left) and dependency tree-ordered context (right). Importance is determined by following a chain of max-pooling decisions over convolutional filters. The weight of arrows shows the relative importance of each token set. Shadow areas show dependency trees, and bold texts are used by convolutional filters.}
    \label{fig:attn-egg-quest}
\end{figure*}

From Figure \ref{fig:egg-quest} and \ref{fig:troll-quest}, a question arises naturally: Why does max-pooling lead to a more stable and better result than mean-pooling?

We find out that the max-pooling layer can be thought of as a kind of attention mechanism \cite{D15-1166}.
With the max-pooling DQN, we can trace back through actions to see which part of trajectories affect the final decision most. An example of attention tokens is shown in Figure \ref{fig:attn-egg-quest} (left part) with the egg quest. The bold texts are the top-3 important attention word-blocks used to make the decision of choosing each action.

\subsection{Zork results}\label{sec:pes}

\begin{figure}
    \centering
    \includegraphics[width=0.45\textwidth]{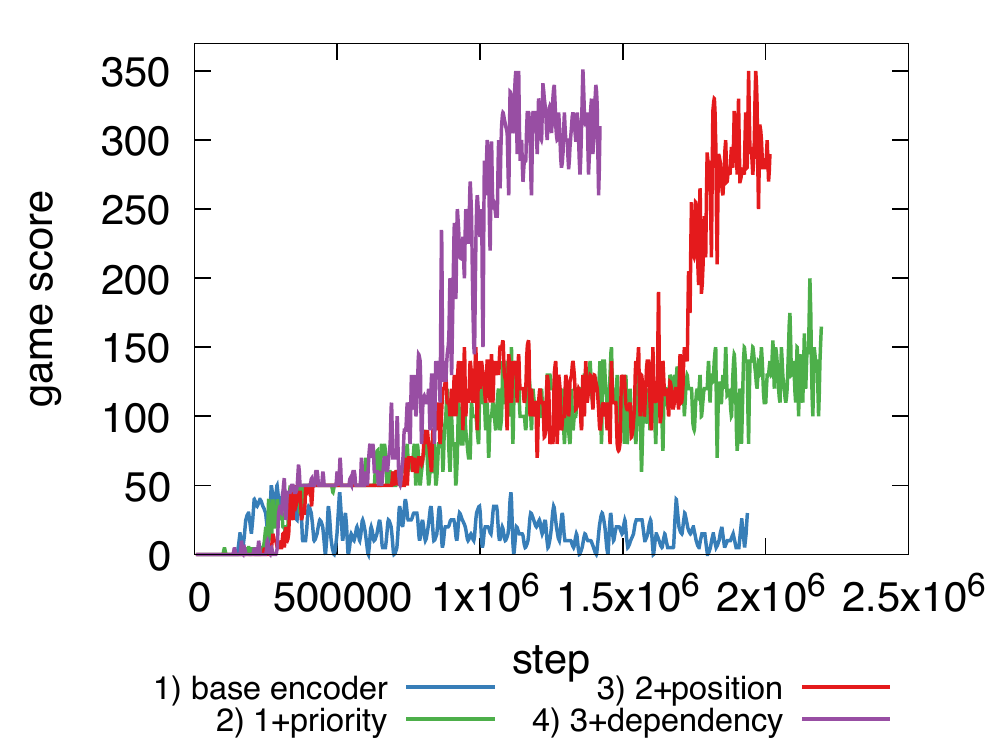}
    \caption{Evaluation results on Zork. The blue curve shows the result of using our base encoder: max-pooling CNN-DQN with weighted sampling. With the use of priority experience sampling we get the green curve. The red curve further improves the performance over priority experience by adding position embeddings. Finally, we use dependency parser reordering on top of priority experience to get the purple curve. The training process of two million steps of base encoder finishes in 22 hours.}
    \label{fig:whole-zork}
\end{figure}

Based on our experience with the egg quest and the troll quest, we choose to use the max-pooling CNN-DQN framework to train agents on the complete Zork. Results are shown in Figure \ref{fig:whole-zork}. Our base encoder (blue) shows bad results for Zork. In 2,000,000 steps of training, there is no evaluation result higher than 50 points, which means for each run, at best we score 5 points for finishing the egg quest. The evaluation result of the blue curve shows that our agent cannot explore Zork beyond the egg quest part.

Zork is a multi-quest game, which means there are multiple distinct trajectories needed to score points. It is important to use priority experience sampling to train an agent on Zork, otherwise the agent tends to converge to one of the local maxima---the egg quest part in our experiment---instead of exploring more rooms of the game, as shown in Figure \ref{fig:whole-zork} comparing the blue curve with the green one.

Furthermore, we apply position embeddings on trajectories as described when playing the egg quest and the troll quest, yielding the red curve with a stable evaluation result of 40 points for a single episode.

\subsection{Dependency parser reordering}
\label{sec:dep}

The CNN encoder, though running a magnitude order faster than the LSTM, encodes local blocks of tokens, while the LSTM encodes a whole sentence. The fact that local blocks are likely to be related is ubiquitous in the domain of image process, but not in natural language processing. In fact, a token and its syntactic dependencies (e.g. a verb and its descriptive adverbs) could appear many words apart in a sentence. Consider the example sentence: {\it you are facing the north side of a white house}. The subject of the sentence could be (notice the bold tokens) \textit{\textbf{you} \textbf{are} \textbf{facing} the north \textbf{side} of a white house}. The token {\it side} is far from other main tokens.

Instead of using convolutional filters on trajectories directly, we use the Stanford dependency parser \cite{chen-manning-2014-fast} to reorder each trajectory so that related tokens are right beside each other. Dependency parsers rearrange tokens of a sentence into a tree, with the subject token as the root (R), and its modifier tokens as children (C). Children of the root can themselves be roots that have children, and so on, recursively. Take the sentence {\it you are facing the north side of a white house} as an example, three subtrees would be generated as 

\begin{enumerate}
    \item \textit{{\bf facing} (R) you (C) are (C) {\bf side} (C)};
    \item \textit{{\bf side} (R) the (C) north (C) {\bf house} (C)};
    \item \textit{{\bf house} (R) of (C) a (C) white (C)}.
\end{enumerate}

Reading the subtrees in a breadth-first way, the three subtrees result in three reordered sub-sentences: \textit{facing you are side}, \textit{side the north house}, and \textit{house of a white}. To avoid size-$N$ convolutional filters striding across boundaries, we add $N-1$ padding tokens among them. E.g. with padding token {\it ``O''} for size-3 convolutional filters, we add two padding tokens:

\textit{facing you are side O O side the north house O O house of a white}

In this way, size-3 convolutional filters cannot stride across each meaningful block.

The result of using a dependency parser reordering is shown in Figure~\ref{fig:whole-zork} (purple curve). With dependency parser reordering, the trained agent can converge in around 1.2 million steps of training, which is faster by half a million steps than the red curve.

\subsection{Repeated bad tries}

We observe that agents tend to repeat themselves, as has also been observed in natural language dialogue generation work \cite{journals/corr/LiMRGGJ16}. Agents tend to `get stuck in a place' and repeat the same action, e.g. \textit{``go west $|$ you need a machete to go west $|$ go west $|$ you need a machete to go west $|$ go west $|$ you need a machete to go west,''} and so on. These repeated tries with no positive reward we call \textit{repeated bad tries}. We determine that trained agents tend to get stuck in areas without short- or long-term positive rewards from those states.

In our experiment with Zork, we find out that out of 2,075,356 training steps, there are 181,209  (8.73\%) repeated bad tries. This behavior gives us the intuition to add an accumulative negative reward on repeated bad samples as a corrective method. In our experiments, we add a negative reward $-0.1$ if we see a repeated sample with a negative reward, and we accumulate the penalty if we immediately see the bad sample again.

\begin{figure}
\centering
\begin{subfigure}[b]{0.45\textwidth}
   \includegraphics[width=1\linewidth]{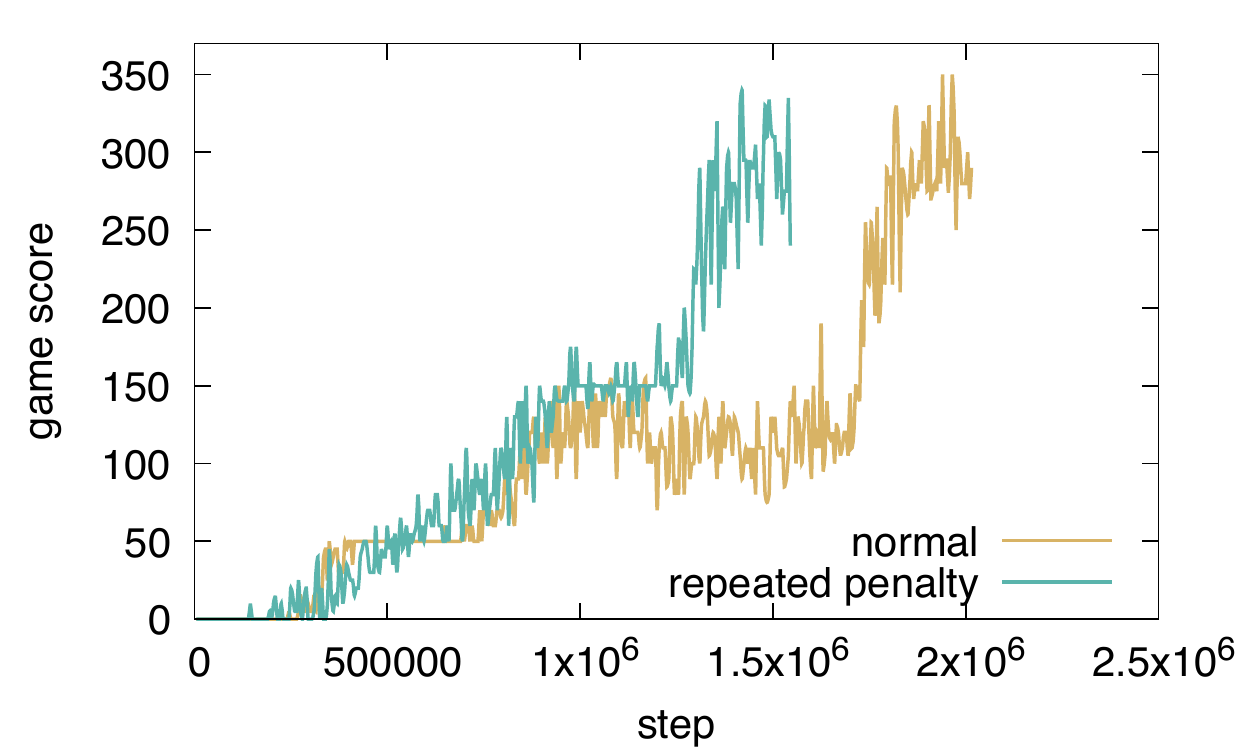}
\end{subfigure}

\begin{subfigure}[b]{0.45\textwidth}
   \includegraphics[width=1\linewidth]{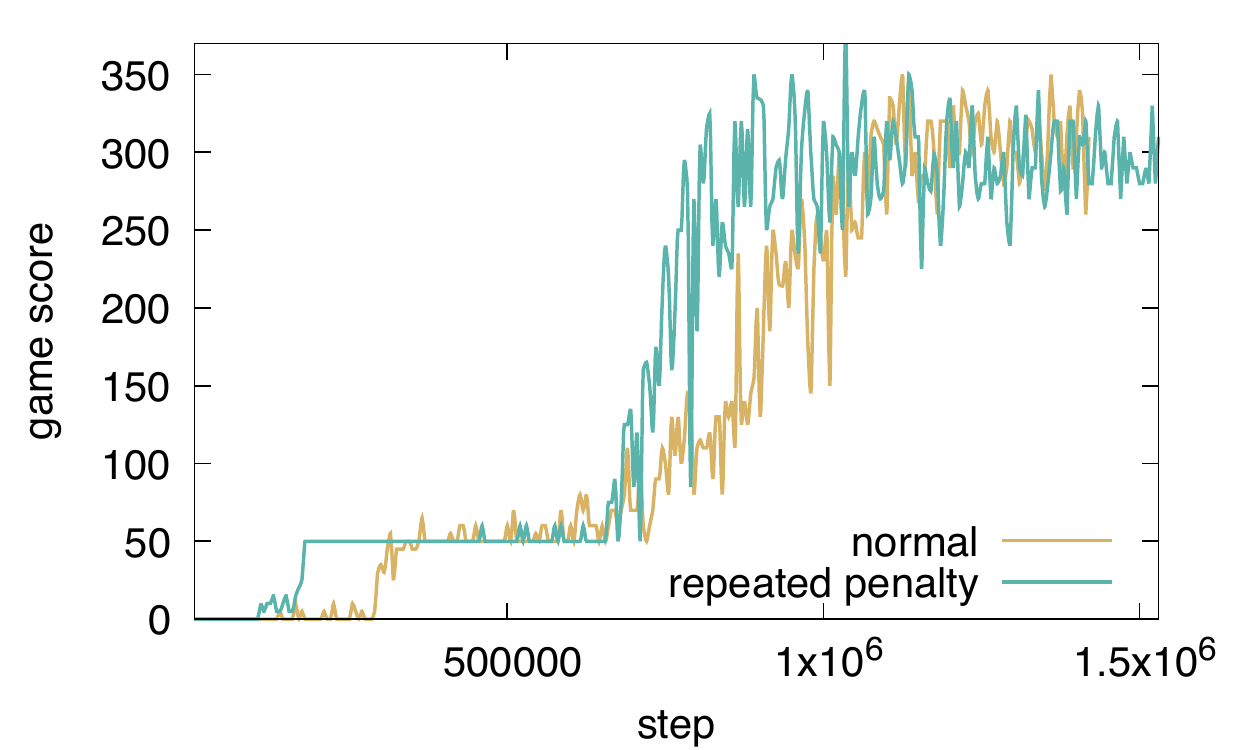}
\end{subfigure}

\caption[Two numerical solutions]{Evaluation results of training CNN-based DQN with repeated penalty (green) VS without repeated penalty (yellow). Both figures use a max-pooling CNN-DQN with position embeddings, however the bottom figure also includes dependency-based reordering (Section~\ref{sec:dep}).}
\label{fig:repeated-penalty}
\end{figure}

\begin{figure}
    \centering
    \includegraphics[width=0.45\textwidth]{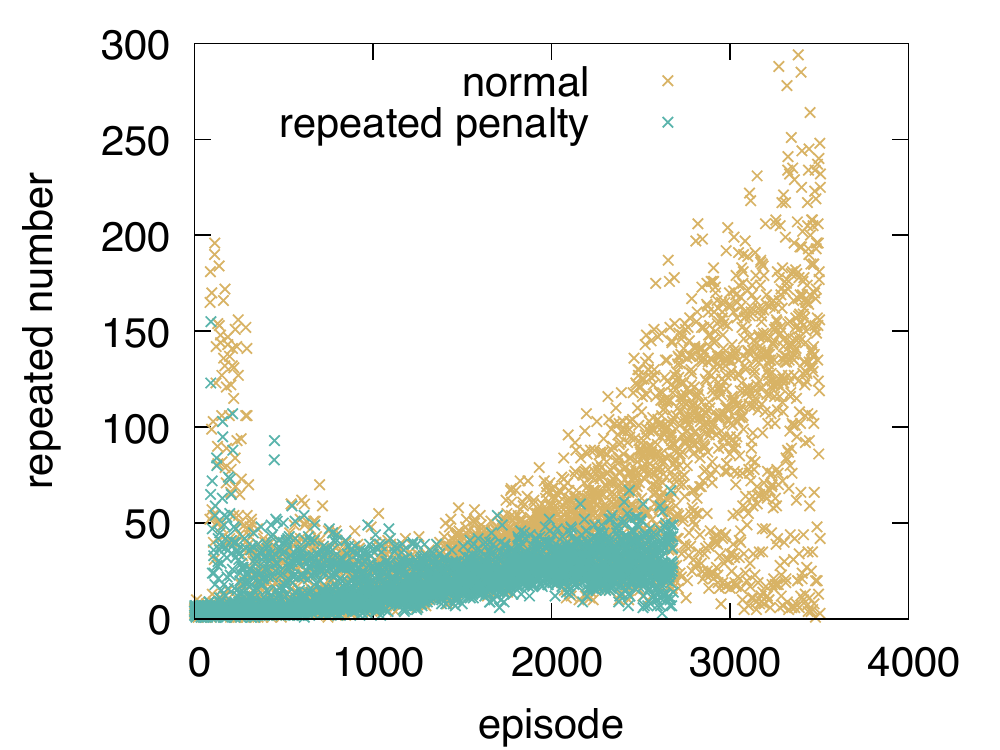}
    \caption{Repeated bad tries in the training process, comparing no repeated penalty (yellow) VS with repeated penalty (green). The x-axis is episode of training, the y-axis is the number of repeated bad tries appearing in each episode. The model used is a max-pooling CNN-DQN with position embeddings.}
    \label{fig:dist-repeated-penalty}
\end{figure}

With the repeated penalty, we can reduce the number of repetitions to 3.51\%; out of 1,596,224 training steps, there are 56,058 repeated bad tries. Compared to using repeated penalty with the previous method (Figure \ref{fig:repeated-penalty}, top, yellow curve), the agent trained with repeated penalty converges much faster, in about half a million training steps. The distribution of repeated bad tries in each training process is plotted in Figure \ref{fig:dist-repeated-penalty}. We also compare using repeated penalty with dependency parser reordering, see Figure \ref{fig:repeated-penalty} (bottom). The benefit is additive; incorporating a repeated penalty can also make the convergence with dependency parser reordering faster, in a quarter million steps.

\section{Related Work}


Several works \cite{D15-1001,P16-1153,journals/corr/LiMRGGJ16,DBLP:journals/corr/abs-1805-07274,fulda2017affordance,textworld-a-learning-environment-for-text-based-games,DBLP:conf/cig/KostkaKKR17} also build agents for text-based games based on the DQN approach designed for action video games \cite{google-atari}.
One key consideration when learning to play text-based games is how to represent game states. Instead of using trajectories, \cite{fulda2017affordance,DBLP:conf/nips/ZahavyHMMM18} use different methods to represent states. Some games allow the use of the special actions \textit{look} and \textit{inventory} to describe the current environment and the player's belongings, and use the combination of the two instead of the trajectory as states.
Our method is more generalized, and avoids the use of \textit{look} and {\it inventory} at every step, which are extra steps that, in certain games (e.g. games with fighting), could lead to a dead state.

Text-based games have a much larger action space to explore than video games of the type evaluated previously \cite{google-atari}, which means that the naive application of the DQN leads to slow or even failing convergence. To reduce the action space, action elimination methods that use both reinforcement learning and NLP-related motivation have been applied. \cite{DBLP:conf/nips/ZahavyHMMM18} use action elimination DQN framework with mathematical bounds to remove unlikely actions, an orthogonal improvement to ours that could be incorporated in future work. \cite{fulda2017affordance} explore affordance by using Word2Vec \cite{NIPS2013_5021} to generate reasonable actions from words, learning, e.g., that {\it eat apple} is more reasonable than {\it eat wheel}.

However, previous works that attempt to play Zork can only finish a very small portion of the game, far from that achievable by human players. \cite{DBLP:conf/nips/ZahavyHMMM18} use the max-pooling CNN-DQN but without position embeddings. Our Zork evaluation result is stable at a score of 40 in one million steps, compared to \cite{DBLP:conf/nips/ZahavyHMMM18}, we get a score of 40 without using the action elimination DQN framework and compared to \cite{DBLP:conf/cig/KostkaKKR17}, that use the LSTM-DQN framework without the action elimination method, we have a huge performance gain.
The generalized method of reward shaping is important for games with multiple sub-quests. \cite{DBLP:journals/corr/abs-1810-12894} use random network distillation to change the instant reward and get improved results on several hard Atari games that require extra exploration.

\section{Conclusion}

By analyzing how the contexts of text-based games are used by our learning approaches, we find that agents make better decisions when they can learn to pay attention to a comprehensible chunk of context. The CNN-DQN with a max-pooling layer is the right tool to reveal the attention information, combining flexibility and speed. However, since linear context is not always a logical chunk of information, dependency parsing gives the CNN the ability to attend focus on syntactically valid chunks.  Our trained agent on Zork can reach the state-of-the-art scores in 10 hours of training.

\bibliography{cog}
\bibliographystyle{IEEEtran}

\end{document}